%% file: main.tex
\documentclass[10pt,twocolumn,letterpaper]{article}

\usepackage[pagenumbers]{cvpr} 
\usepackage{adjustbox}
\usepackage{arydshln}
\usepackage{pdflscape}
\usepackage[accsupp]{axessibility}

\input{preamble}
\definecolor{cvprblue}{rgb}{0.21,0.49,0.74}
\usepackage[pagebackref,breaklinks,colorlinks,allcolors=cvprblue]{hyperref}

\newcommand\blfootnote[1]{%
  \begingroup
  \renewcommand\thefootnote{}\footnote{#1}%
  \addtocounter{footnote}{-1}%
  \endgroup
}

\def\paperID{*****} 
\def\confName{NTIRE}
\def\confYear{2026}
\input{sec/authors}

\title{Robust Deepfake Detection, NTIRE 2026 Challenge: Report}

\begin{document}
\maketitle
\blfootnote{$^\dagger$ B. Hopf (\href{mailto:benedikt.hopf@uni-wuerzburg.de}{benedikt.hopf@uni-wuerzburg.de}) and R. Timofte (\href{mailto:radu.timofte@uni-wuerzburg.de}{radu.timofte@uni-wuerzburg.de}) from the University of W\"urzburg, Germany, were the challenge organizers, while the other authors participated in the challenge.
See \cref{sec:teams} for team details.\\
}
\input{sec/0_abstract}    
\input{sec/1_intro}

\input{sec/2_details}
\input{sec/3_methods}
\input{sec/4_conclusion}

{
    \small
    \bibliographystyle{ieeenat_fullname}
    \bibliography{main}
}

\input{sec/X_suppl}

\end{document}

%% file: sec/authors.tex
\author{
Benedikt Hopf$^\dagger$ \and
Radu Timofte$^\dagger$ \and
Chenfan Qu \and 
Junchi Li \and
Fei Wu \and
Dagong Lu \and
Mufeng Yao \and
Xinlei Xu \and
Fengjun Guo \and
Yongwei Tang \and
Zhiqiang Yang \and
Zhiqiang Wu \and 
Jia Wen Seow \and
Hong Vin Koay \and
Haodong Ren \and
Feng Xu \and 
Shuai Chen \and
Minh-Khoa Le-Phan \and
Minh-Hoang Le \and
Trong-Le Do \and
Minh-Triet Tran \and
Chih-Yu Jian \and
Yi-Fan Wang \and
Bang-Kang Chen \and
You-Chen Chao \and
Chia-Ming Lee \and
Fu-En Yang \and
Yu-Chiang Frank Wang \and
Chih-Chung Hsu \and
Aashish Negi \and 
Hardik Sharma \and 
Prateek Shaily \and 
Jayant Kumar \and 
Sachin Chaudhary \and 
Akshay Dudhane \and 
Praful Hambarde \and 
Amit Shukla \and 
Jielun Peng \and 
Yabin Wang \and 
Yaqi Li \and 
Jincheng Liu \and 
Xiaopeng Hong \and 
Krish Wadhwani \and
Liam Fitzpatrick \and
Utkarsh Tiwari \and 
Bilel Benjdira \and 
Anas M. Ali \and 
Wadii Boulila \and 
Cristian Lazo Quispe \and 
Aishwarya A \and
Akshara S \and
Ashwathi N \and
Jiachen Tu \and
Guoyi Xu \and
Yaoxin Jiang \and
Jiajia Liu \and
Yaokun Shi
}

%% file: sec/0_abstract.tex
\begin{abstract}

Robustness is a long-overlooked problem in deepfake detection. However, detection performance is nearly worthless in the real world if it suffers under exposure to even slight image degradation. In addition to weaker degradations that can accidentally occur in the image processing pipeline, there is another risk of malicious deepfakes that specifically introduce degradations, purposefully exploiting the detector's weaknesses in that regard. Here, we present an overview of the NTIRE 2026 Robust Deepfake Detection Challenge, which specifically addresses that problem. Participants were tasked with building a detector that would later be tested on an unknown test-set, which included both common and uncommon degradations of various strengths. With a total number of 337 participants and 57 submissions to the final leaderboard, the first edition of the challenge was well received. To ensure the reliability of the results, participants were given only 24h to complete the test run with no labels provided, limiting the possibility of training on the test data. Furthermore, the top solutions were scored on a private test-set to detect any such overfitting. This report presents the competition setting, dataset preparation, as well as details and performance of methods. 
Top methods rely on large foundation models, ensembles, and degradation training to combine generality and robustness.
\end{abstract}

%% file: sec/1_intro.tex
\section{Introduction}

Deepfake detection has significantly increased in importance over the last few years, as generated content on the internet has become more prominent. Although recent literature has improved generality towards dataset and method shifts~\cite{Shiohara2022SBI, Yan2023UCFUC, Yan2023LSDA, Li2019FaceXRay, yan2025orthogonalsubspacedecompositiongeneralizable, Zhuang2022UIAViTUI, Larue2022SeeABLESD, nguyen2024laa}, the aspect of robustness to low-quality images has not been studied as deeply. Some methods~\cite{nguyen2024laa, lu2022ddrc, Yan2023DeepfakeBenchAC} studied the performance implications of low image-quality images, and~\cite{hopf2025practicalmanipulationmodelrobust} showed that modeling degradations can improve performance even on benchmark datasets. Following this line of work, we propose this challenge to engage the community in developing models that are not restricted to high-quality deepfakes, but can also work on lower-quality hard samples.

This is particularly important because~\cite{hopf2025practicalmanipulationmodelrobust} showed that such image degradations could be maliciously used to circumvent detection, putting the entire usefulness of a deepfake detector at risk if it can be easily circumvented.

This challenge is one of the challenges associated with the NTIRE 2026 Workshop~\footnote{\url{https://www.cvlai.net/ntire/2026/}} on:
deepfake detection~\cite{ntire26deepfake}, 
high-resolution depth~\cite{ntire26hrdepth},
multi-exposure image fusion~\cite{ntire26raim_fusion}, 
AI flash portrait~\cite{ntire26raim_portrait}, 
professional image quality assessment~\cite{ntire26raim_piqa},
light field super-resolution~\cite{ntire26lightsr},
3D content super-resolution~\cite{ntire263dsr},
bitstream-corrupted video restoration~\cite{ntire26videores},
X-AIGC quality assessment~\cite{ntire26XAIGCqa},
shadow removal~\cite{ntire26shadow},
ambient lighting normalization~\cite{ntire26lightnorm},
controllable Bokeh rendering~\cite{ntire26bokeh},
rip current detection and segmentation~\cite{ntire26ripdetseg},
low light image enhancement~\cite{ntire26llie},
high FPS video frame interpolation~\cite{ntire26highfps},
Night-time dehazing~\cite{ntire26nthaze,ntire26nthaze_rep},
learned ISP with unpaired data~\cite{ntire26isp},
short-form UGC video restoration~\cite{ntire26ugcvideo},
raindrop removal for dual-focused images~\cite{ntire26dual_focus},
image super-resolution (x4)~\cite{ntire26srx4},
photography retouching transfer~\cite{ntire26retouching},
mobile real-word super-resolution~\cite{ntire26rwsr},
remote sensing infrared super-resolution~\cite{ntire26rsirsr},
AI-Generated image detection~\cite{ntire26aigendet},
cross-domain few-shot object detection~\cite{ntire26cdfsod},
financial receipt restoration and reasoning~\cite{ntire26finrec},
real-world face restoration~\cite{ntire26faceres},
reflection removal~\cite{ntire26reflection},
anomaly detection of face enhancement~\cite{ntire26anomalydet},
video saliency prediction~\cite{ntire26videosal},
efficient super-resolution~\cite{ntire26effsr},
3d restoration and reconstruction in adverse conditions~\cite{ntire26realx3d},
image denoising~\cite{ntire26denoising},
blind computational aberration correction~\cite{ntire26aberration},
event-based image deblurring~\cite{ntire26eventblurr},
efficient burst HDR and restoration~\cite{ntire26bursthdr},
low-light enhancement: `twilight cowboy'~\cite{ntire26twilight},
and efficient low light image enhancement~\cite{ntire26effllie}.

%% file: sec/2_details.tex
\section{Challenge Details}

The NTIRE 2026 Robust Deepfake Detection Challenge consisted of two phases. First, in the training and validation phase, participants were provided with a small training and validation set of 1000 and 100 images, respectively. Participants were allowed to use any additional public datasets for training to improve robustness. The provided training dataset had the main purpose of providing a sample of what degradations to expect. The validation dataset, however, was not intended to be used for training. Validation was, therefore, performed through Codabench~\cite{Xu_2022}, without access to ground truth labels. We do have to note, however, that we cannot fully guarantee that participants did not manually label the dataset. Given that caveat, we made the validation dataset small, such that the additional effect for training should have been limited.

Testing was performed on a public test set, which was released 24h before the end of the test phase, in order to make it harder to do any further finetuning on the test set. Furthermore, only one submission to the test server was allowed during that phase to also prevent validation using the test set. Again, despite these security measures, we cannot fully rule out the possibility that the test set could have been manually annotated and used for training and/or validation.

Therefore, we employed another layer of security for the top submissions: All submissions provided their code and pretrained models, so we could evaluate the top contenders on another, fully unknown, private test-set. This fully rules out the possibility of any finetuning on the test-set. For the top methods, which were ranked on the private test set, private test performance takes precedence over the public performance. We are happy to state that we did not observe major shifts in ranking between the test sets, so we do not suspect any finetuning on the test-set. Final results are shown in \cref{tab:results}.

\begin{table}
    \centering
    \begin{tabular}{c:c|cc} \toprule
        Rank & Team name & Public Test & Private Test \\ \midrule
        1 & ShallowReal &  \underline{0.9218} &  \textbf{0.9168} \\
        2 & INTSIG & 0.8901 &  \underline{0.8824} \\
        3 & AntInternational & \textbf{0.9234} & 0.8691 \\
        4 & HCMUS-Aqua & 0.8775 &  0.8523 \\ 
        5 & acvlab & 0.8461 & 0.8007 \\
        6 & Reagvis Lab & 0.8430 & 0.7934 \\
        7 & Hit Virlab & 0.8418 & 0.7620 \\
        8 & \textit{Anonymous} &  0.839 & - \\
        9 & Zeke & 0.824 & - \\
        10 & TCD Vision &  0.8045  & - \\
        11 & PSU &  0.7056 & - \\
        12 & AI4Good & 0.6865 & - \\
        13 & Acube & 0.6860 & - \\
        14 & NTR & 0.5999 & - \\

        \bottomrule
    \end{tabular}
    \caption{\textbf{Results of the challenge.} The top submissions have additionally been scored on a private test set to verify generality.}
    \label{tab:results}
\end{table}

\subsection{Dataset}

Due to the multi-phase design of the challenge, we provided four datasets. All images are based on CelebV-HQ~\cite{zhu2022celebvhq}, but use different degradations and fake methods. We generally attempt to have the splits be comparable, but require slightly more generalization ability for the validation and testing splits in order for them to be somewhat out-of-domain.

All images are crops of faces with some margin around them. Furthermore, only face swapping or reenactment methods were used, so no fully-synthetic images.

\vspace{2mm}\noindent\textbf{Training split}
The training split consists of 1000 images, equally split between real and fake. Fake images are created using the baseline method FaceSwap~\cite{faceswap}. For generalizability, participants were able to use additional public datasets.

The degradation model is a slightly changed version of the baseline paper PMM~\cite{hopf2025practicalmanipulationmodelrobust}, using different strengths of Gaussian noise, JPEG compression, smoothing, resizing, and color and contrast adjustments. This training set provides basic but relatively strong degradations for training. We specifically did not include uncommon degradations in order to require models to generalize and to examine the hypothesis that a sufficiently diverse degradation model can also work well on unseen degradations.

\vspace{2mm}\noindent\textbf{Validation split}
The validation split consists of 100 images, equally split between real and fake. To have a combination of in-dataset and generalization performance, the validation set is an extension of the training set: In addition to FaceSwap~\cite{faceswap} it also uses StyleFeatureEditor~\cite{bobkov2024ganediting} for a second fake method. This method has already been used in the baseline paper~\cite{hopf2025practicalmanipulationmodelrobust}, so it was not fully unknown to the participants.

Similarly, the validation split uses the same degradations as the training split, but adds speckle and Poisson noise. These are still degradations used in~\cite{hopf2025practicalmanipulationmodelrobust}, so we expected participants to be aware of them, so they would not have been good fits for the test sets. Still, they provide a difference from the training dataset, allowing validation.

\vspace{2mm}\noindent\textbf{Public test split}
The public test split consists of 1000 images, equally split between real and fake. We extend the scheme from before by adding to the fake methods. This has the rationale that while we mainly focus on robustness to degradations, we also want to include some amount of cross-generator testing. In addition to the previous methods, we use FSGAN~\cite{nirkin2019fsgan}. This method has not been used in the baseline paper, so just using the baseline would not have directly targeted the test set. 

For the degradation model, we again use all common degradations (noise, blur, compression, ...) but with slightly varying parameters compared to the training and validation sets. Furthermore, we add some less common degradations in the form of salt-and-pepper noise, black-and-white color filters, and overlay addition. The overlay is added by taking the same picture, scaling it up by a factor of 2 to 4, and adding it to the image using a random transparency value of $[0, 0.33]$. 

\vspace{2mm}\noindent\textbf{Private test split}
The private test split consists of 1000 images, equally split between real and fake. In addition to the public split, it adds another fake method and two more uncommon image degradations. This split was only used to validate the top solutions against overfitting to the public test set.

\subsection{Evluation metric}

As the evaluation metric, we follow the common protocol in face forgery detection~\cite{Shiohara2022SBI, nguyen2024laa, hopf2025practicalmanipulationmodelrobust, Yan2023DeepfakeBenchAC} and use the area under the receiver operating characteristic curve (AUC). This metric has the advantage over accuracy, that it is theshold-free, avoiding problems with calibration on unseen test sets.

%% file: sec/3_methods.tex
\section{Methods}

The following sections have been written by the individual teams (except for minor changes to figure references and captions to be distinct in the overview paper, as well as adding inline citations, and grammatical reformulations to better fit the report style \eg changing ``we'' to ``they''). All figures are additionally provided in a double-column layout in the supplementary.

\subsection{SHALLOW REAL: DINO-MAC}
\input{sec/methods/shallowreal_description}

\subsection{INTSIG: LOGER: Local-Global Ensemble for Robust Deepfake Detection in the Wild}
\input{sec/methods/intsig_description}

\subsection{ANT INTERNATIONAL: An Ensemble of Architecturally-Diverse Large-Scale Vision}
\input{sec/methods/antint_description}

\subsection{HCMUS-AQUA: Robust Deepfake Detection via Multi-Stream DINO-CLIP Fusion and Discretized Voting}
\input{sec/methods/hcmus_description}

\subsection{ACVLAB: Select and Detect: Quality-Aware Expert Routing and Robust Optimization for Deepfake Detection}
\input{sec/methods/acvlab_description}

\subsection{REAGVIS LABS: Beyond Backbones: Degradation-Aware Prototype Fusion for Robust Deepfake Detection}
\input{sec/methods/reagvis_description}

\subsection{HIT-VIRLAB: Hierarchical Adaptive Feature Aggregation with Degraded-Original Consistency Learning for Robust Deepfake Detection}
\input{sec/methods/hit_description}

\subsection{Anonymous: LoRA Fine-Tuning for CLIP}
\input{sec/methods/anonymous1_description}

\subsection{ZEKE: Robust Deepfake Detection using Large Scale Vision Transformers}
\input{sec/methods/zeke_description}

\subsection{TCD VISION: Robust Deepfake Detection with Parameter-Efficient CLIP Fine-Tuning and PMM-Style Degradation Augmentation}
\input{sec/methods/tcd_description}

\subsection{PSU: PRISM: Paradigm-diverse Representation Integration for Synthesis-artifact Manifold Detection}
\input{sec/methods/psu_description}

\subsection{AI4GOOD: Self-Supervised Adversarial Training for Robust Deepfake Detection}
\input{sec/methods/ai4good_description}

\subsection{ACUBE: Robust Deepfake Detection using ConvNeXt with Frequency-Aware Fusion and Regularized Training Strategy}
\input{sec/methods/acube_description}

\subsection{NTR: DINOv3 ViT-B/16 Linear Probe Ensemble for Robust Deepfake Detection}
\input{sec/methods/ntr_description}

%% file: sec/methods/shallowreal_description.tex
\begin{figure}
    \centering
    \includegraphics[width=1.0\linewidth]{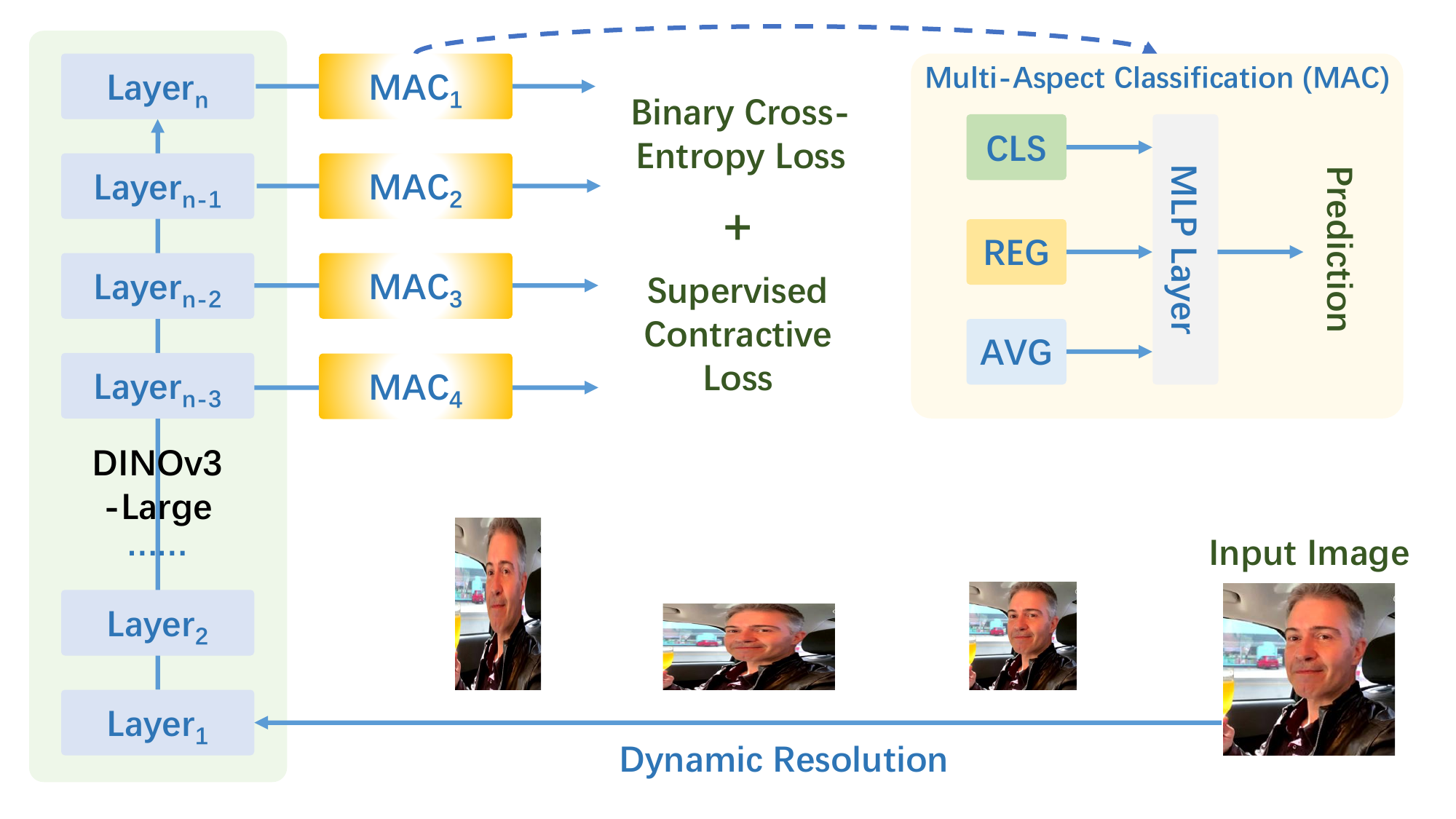}
    \caption{Overall pipeline of ShalloReal's \textit{DINO-MAC} model.}
    \label{fig:shallowreal}
\end{figure}

As illustrated in Figure~\ref{fig:shallowreal}, they frame the task as a binary classification problem~\cite{dinomac}. Their model employs the DINOv3-Large~\cite{simeoni2025dinov3} architecture as its backbone, which is fine-tuned using Low-Rank Adaptation (LoRA)~\cite{Hu2021LoRALA} with a rank of 32 and an alpha of 64.

The final prediction is generated by a Multi-Aspect Classification (MAC) head that processes features from the DINOv3 backbone. This module aggregates information from multiple sources: the [CLS] token, four [REG] register tokens, and an [AVG] token representing the average of the other patch tokens. These six 1024-dimensional feature vectors are concatenated into a single 6144-dimensional tensor. This tensor is then fed into a Multi-Layer Perceptron (MLP)—composed of two fully-connected layers with an intermediate ReLU activation—to produce the final classification score.

During training, they implemented four key techniques:

1. Dynamic Resolution: Input images were resized to a random resolution where both height and width were sampled from the range [384, 1152] (multiples of the patch size 16), using a randomly selected interpolation method.

2. Deep Supervision: Auxiliary classification heads were attached to the last four layers of the DINOv3 backbone. A loss was computed for each, but only the output from the final layer was used during inference. A Dropout rate of 0.2 was used in the fully-connected layer of each head.

3. Metric Learning: A Supervised Contrastive Learning (SCL) loss was added as an auxiliary objective, complementing the primary binary cross-entropy (BCE) loss to enhance feature discrimination.

4. Stochastic Depth: A random drop path rate of 0.1 is applied.

%% file: sec/methods/intsig_description.tex
\begin{figure*}
    \centering
    \includegraphics[width=0.7\linewidth]{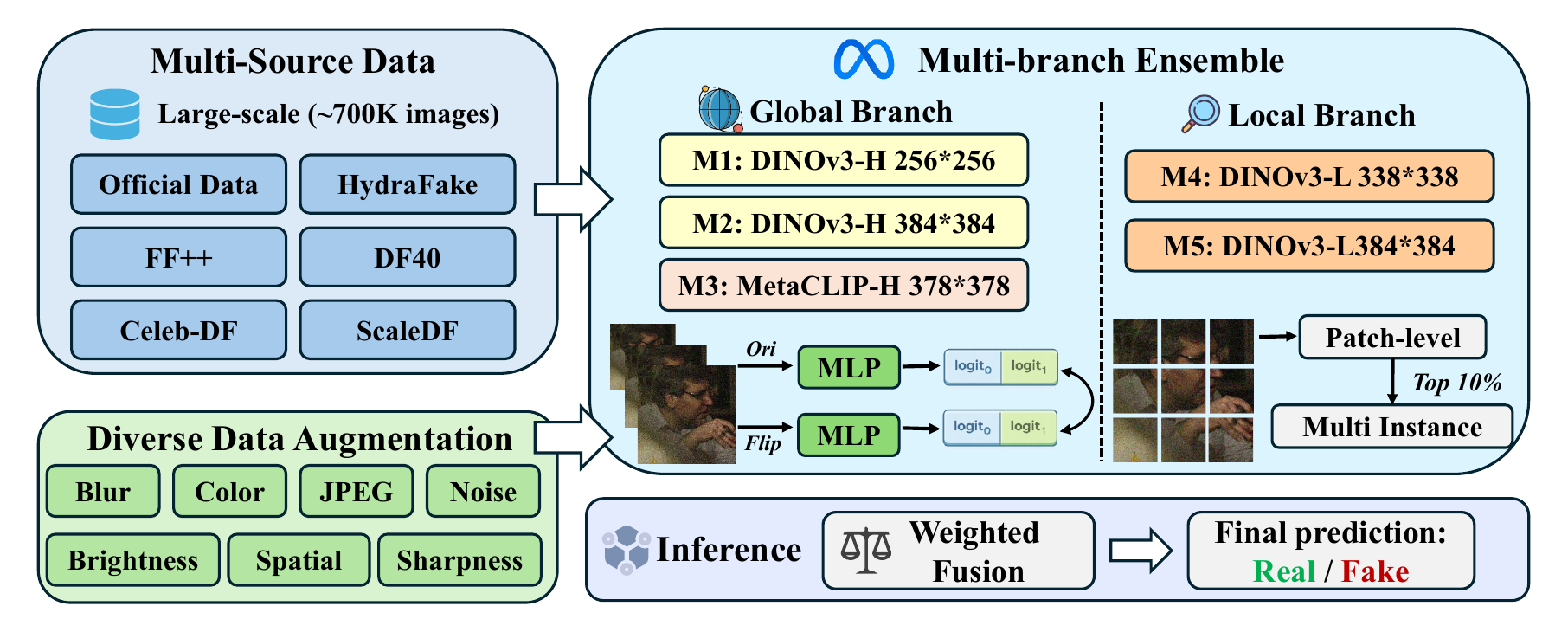}
    \caption{Overall pipeline of INTSIG's \textit{LOGER: Local-Global Ensemble for Robust Deepfake Detection in the Wild}.}
    \label{fig:intsig}
\end{figure*}

Team INTSIG proposes LOGER, a local--global ensemble framework for robust deepfake detection, designed to jointly capture global semantic inconsistencies and localized forensic artifacts (\cref{fig:intsig}).

The \textbf{global branch} performs full-image detection at multiple scales with diverse backbones. They employ three models: M1 and M2 share a DINOv3-Huge~\cite{simeoni2025dinov3} backbone with full-parameter fine-tuning and a two-layer MLP classification head. M1 is trained and inferred at \textbf{256$\times$256}, while M2 is trained at the same resolution but inferred at \textbf{384$\times$384}, preserving fine-grained forensic details that lower resolutions would discard. Both are trained with Focal Loss~\cite{focalloss}. M3 uses MetaCLIP2-Huge~\cite{chuang2025metaclip2worldwide} to introduce backbone heterogeneity: DINOv3 encodes spatial and structural priors via self-supervised pre-training, whereas MetaCLIP2 provides complementary semantic grounding through contrastive image-text alignment, reducing correlated errors in the ensemble. M3 adopts a staged loss schedule---cross-entropy for the first 20\% of training, then Focal Loss for the remaining 80\%---to ensure stable convergence before shifting focus toward hard samples.

The \textbf{local branch} targets forgery traces concentrated in specific facial regions that global averaging tends to dilute. M4 and M5 use DINOv3-Large~\cite{simeoni2025dinov3} with patch-level modeling: the input is split into non-overlapping patches, each mapped to real/fake logits, and aggregated via Multiple Instance Learning (MIL) top-$k$ pooling that selects only the top \textbf{10\%} highest-response patches. This strategy enhances sensitivity to small forged areas while suppressing noise from normal patches. Training uses a multi-term objective combining image-level cross-entropy, a pairwise AUC surrogate loss, a patch-level MIL loss, and regularization, providing dual-level supervision at both the aggregated image level and individual patch level. M4 is trained at \textbf{224$\times$224} and inferred at \textbf{384$\times$384}; M5 is initialized from M4 and further fine-tuned at \textbf{338$\times$338}, producing two complementary local detectors through continuation training.

During inference, they apply horizontal-flip TTA for M1, M2, M4, and M5. For final fusion, they convert each model's output into a single directional evidence score $l_{\text{fake}} - l_{\text{real}}$, average the five scores with uniform weights, and apply a sigmoid function. Fusing in the logit space before the sigmoid retains each model's full confidence range, yielding more robust predictions than probability averaging.

%% file: sec/methods/antint_description.tex
First, they conducted experiments on several state-of-the-art image foundation models, including DinoV3 \cite{simeoni2025dinov3}, SigLIP \cite{zhai2023siglip}, EVA-giant \cite{evagiant23} and I-JEPA \cite{ijepa23}. The experiments revealed \textbf{DINOv3} backbone demonstrated superior generalization capabilities due to its self-supervised pre-training on the vast and highly diverse LVD-1689m dataset. This endows the model with more universal visual features that are less susceptible to the biases of a specific training set.

\begin{figure*}
    \centering
    \includegraphics[width=0.7\textwidth]{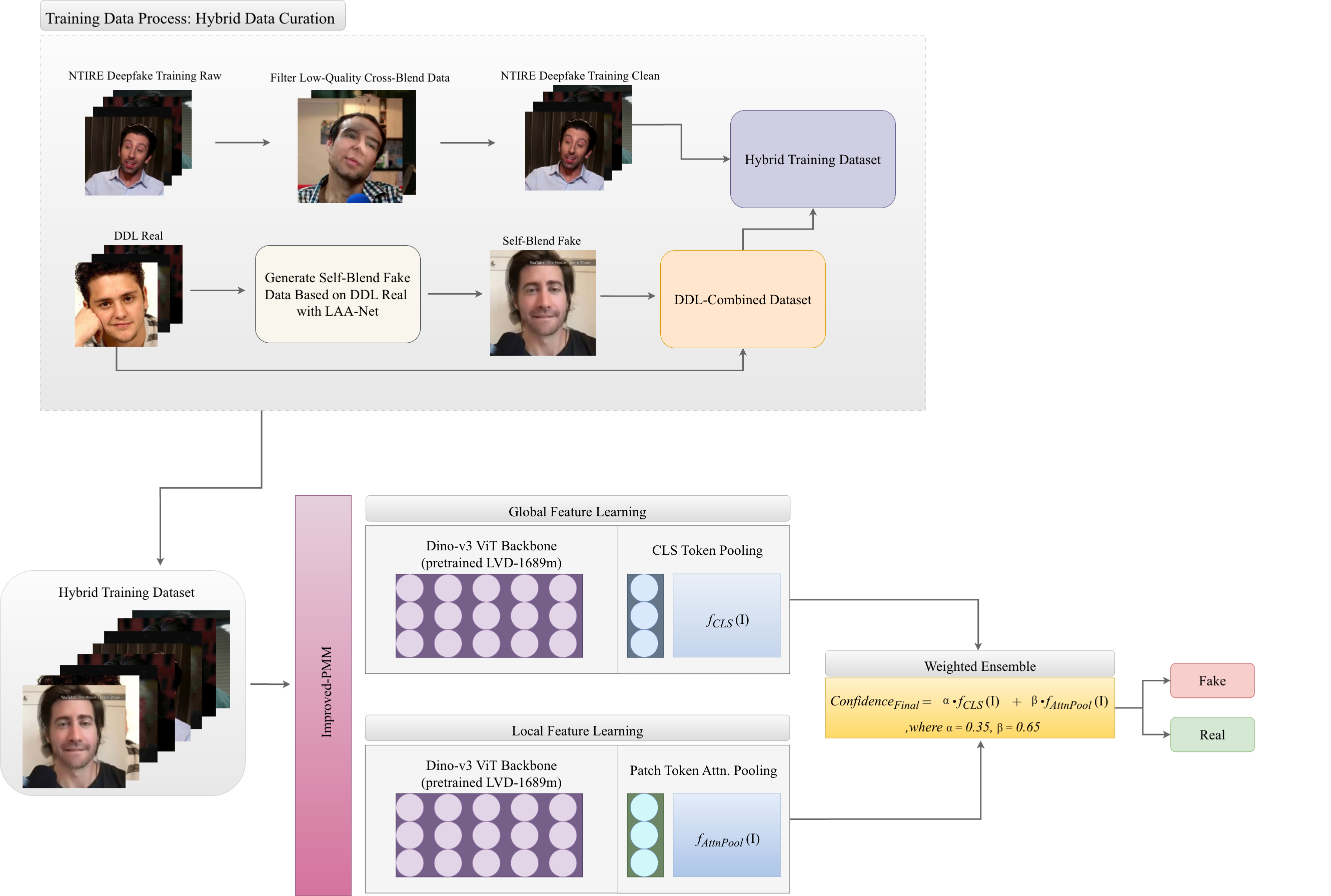}
    \caption{Overall pipeline of AntInternational's \textit{An Ensemble of Architecturally-Diverse Large-Scale Vision Transformers}.}    
    \label{fig:antint}
\end{figure*}

\paragraph{Hybrid Dataset Curation.}
They integrated the DDL dataset\cite{miao2025ddllargescaledatasetsdeepfake} into the curation strategy due to its huge coverage in the implementation of deepfake methods and diverse forgery scenarios to mimic real-world conditions. The final hybrid training set was constructed as follows: \textit{i.}, official data, with low-quality fake images which hindered model convergence, were filtered out, \textit{ii.} real, unmanipulated images from the DDL dataset, and \textit{iii.} self-blended DDL images using LAA-Net \cite{nguyen2024laa}.

\paragraph{Architecturally-Diverse Ensembling.}
They employed an ensemble model derived from the same DINOv3 backbone to maximize feature diversity and mitigate the risk of overfitting to a single architectural bias (\cref{fig:antint}).

\noindent\textbf{ViT-CLS.} By utilizing the standard CLS token output from the DINOv3 backbone. This model was designed to capture the global feature and identify large-scale inconsistencies or artifacts that manifest over broad regions of the image. 

\noindent\textbf{ViT-AttnPool.} This second model was specifically implemented to focus on localized, subtle artifacts that a global-only view might miss. It replaces the standard pooling mechanism with a custom AttentionPooling layer that operates directly on all patch tokens. It learns to focus on specific patches—areas likely to contain subtle artifacts, like seam boundaries or facial distortions. 

The final submission score is a weighted average of the predictions from our two independently
trained models. We determined an optimal 35/65 weighting scheme:

\begin{equation}
    \textit{Confidence}_{\text{final}} = \alpha \cdot f_{\text{CLS}}(I) + \beta \cdot f_{\text{AttnPool}}(I)
    \label{eq:ensemble_ant}
\end{equation}
where the weights \(\alpha=0.35\) and \(\beta=0.65\) were empirically determined.

\paragraph{Face-Aware Augmentation}
Model robustness against real-world image degradations was further enhanced through an \textbf{Improved-PMM-Aug} pipeline. This strategy builds upon the Practical Manipulation Model (PMM) \cite{hopf2025practicalmanipulationmodelrobust} by expanding its suite of degradations to include grayscale conversion, pixelation, and various noise functions (e.g., salt \& pepper, speckle). They integrated a face-aware mechanism for the text-based 'distractor' overlays. This ensures that random text augmentations are not placed over detected facial regions, thereby preventing the model from being distracted and forcing it to learn from the most forensically relevant parts of the image.

%% file: sec/methods/hcmus_description.tex
\begin{figure*}
    \centering
    \includegraphics[width=0.8\linewidth]{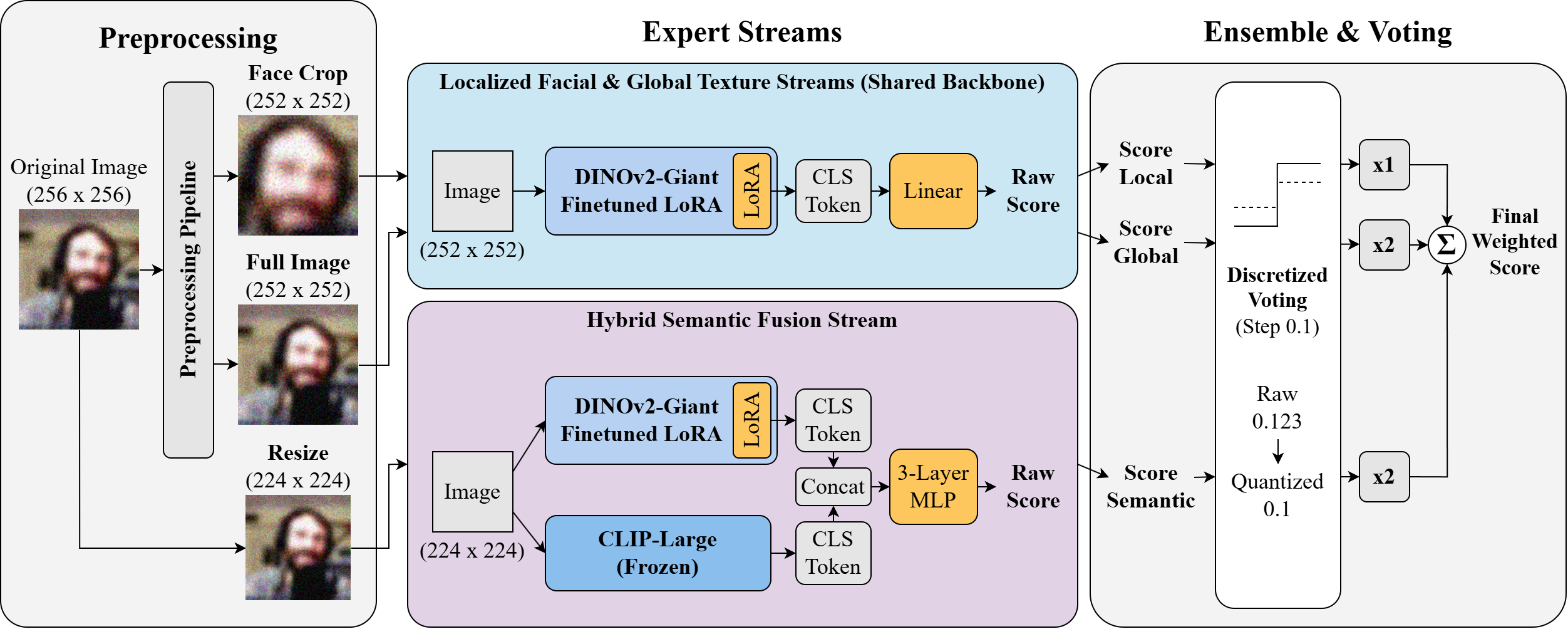}
    \caption{Overall pipeline of HCMUS-Aqua's \textit{Robust Deepfake Detection via Multi-Stream DINO-CLIP Fusion and Discretized Voting}. The system processes inputs through three specialized expert streams. The \textit{Localized Facial} and \textit{Global Texture} streams maintain native signal integrity ($252\times252$) utilizing a shared DINOv2-Giant backbone~\cite{oquab2024dinov2}. The \textit{Hybrid Semantic Fusion} stream ($224\times224$) concatenates geometric features from DINOv2~\cite{oquab2024dinov2} with semantic features from a frozen CLIP-Large model~\cite{radford2021clip}. Trainable components (LoRA modules~\cite{Hu2021LoRALA} and MLPs) are highlighted in red, while frozen/pretrained backbones are depicted in blue. Finally, raw probabilities are quantized to 0.1 precision steps and aggregated via \textbf{Discretized Probability Voting} (using a calibrated 1:2:2 weighting ratio for Local:Global:Fusion) to output a robust final score.}
    \label{fig:hcmus_figure}
\end{figure*}

Their method~\cite{lephan2026ntiredeepfake} addresses robust deepfake detection through a comprehensive foundation-driven framework designed to mitigate spatial attention drift under real-world compound degradations (\cref{fig:hcmus_figure}):
\begin{itemize}
    \item \textbf{Global Texture Stream:} Operating at a native $252\times252$ resolution, this stream sweeps the uncropped image using a LoRA-adapted DINOv2-Giant backbone. It explicitly isolates macro-contextual anomalies, including spatial illumination inconsistencies and mismatched compression profiles between the synthetic face and pristine background.
    \item \textbf{Localized Facial Stream:} This pathway anchors biometric geometry by extracting a $1.3\times$ expanded facial crop, resized to $252\times252$ to prevent interpolation loss. To ensure stability under extreme degradation that breaks standard detectors, it incorporates a robust 7-step cascaded recovery pipeline (utilizing bilateral filtering, GFPGAN enhancement, and CLAHE), which successfully reduces spatial parsing failure rates from 15\% to just 1.8\%.
    \item \textbf{Hybrid Semantic Fusion Stream:} Acting as a semantic safety net, this $224\times224$ stream concatenates geometric representations from DINOv2 with language-supervised priors from a strictly frozen CLIP-Large~\cite{radford2021clip} backbone. This enforces strict semantic verification, allowing the model to detect logical impossibilities (e.g., melting accessories) that evade pure texture analysis.
    \item \textbf{Extreme Compound Degradation Engine:} To neutralize texture shortcut learning, the training pool is subjected to a randomized 18-operation degradation pipeline. By systematically applying cyclic JPEG compression, H.264 packet loss simulation, and optical blur, the backbone is forced to abandon fragile high-frequency cues in favor of invariant structural geometry.
    \item \textbf{Balanced Multi-Domain Optimization:} The team curated a highly balanced master pool of 377,343 frames across 14 diverse datasets (Celeb-DF-v3~\cite{cdfv3}, DeeperForensics-1.0~\cite{deeperforensics10}, HIDF~\cite{hidf}, RedFace~\cite{redface}, DF40~\cite{yan2024df40}, DDL~\cite{miao2025ddllargescaledatasetsdeepfake}, FaceForensics++~\cite{roessler2019faceforensicspp}, Celeb-DF-v2~\cite{Li2019CelebDFv2}, DeepFakeDetection~\cite{dfd}, DFDC~\cite{Dolhansky2020DFDC}, DFDCP~\cite{Dolhansky2019DFDCP}, FFIW~\cite{ffiw}, FaceShifter~\cite{faceshifter}, and UADFV~\cite{uadfv}). Entire Face Synthesis media was explicitly filtered out to tightly isolate face-swapping boundaries and prevent domain memorization.
\end{itemize}
To address prediction instability, the system uses a Discretized Probability Voting mechanism. Raw probabilities from the three streams are quantized into 11 discrete levels (0.1 increments) prior to aggregation. The ensemble applies a 1:2:2 weighting ratio (Local:Global:Fusion). This configuration reduces the relative influence of the localized stream in conditions where noise degrades image blending edges, relying instead on the global and semantic streams to maintain AUC stability.

%% file: sec/methods/acvlab_description.tex
\begin{figure*}
    \centering
    \includegraphics[width=0.85\linewidth]{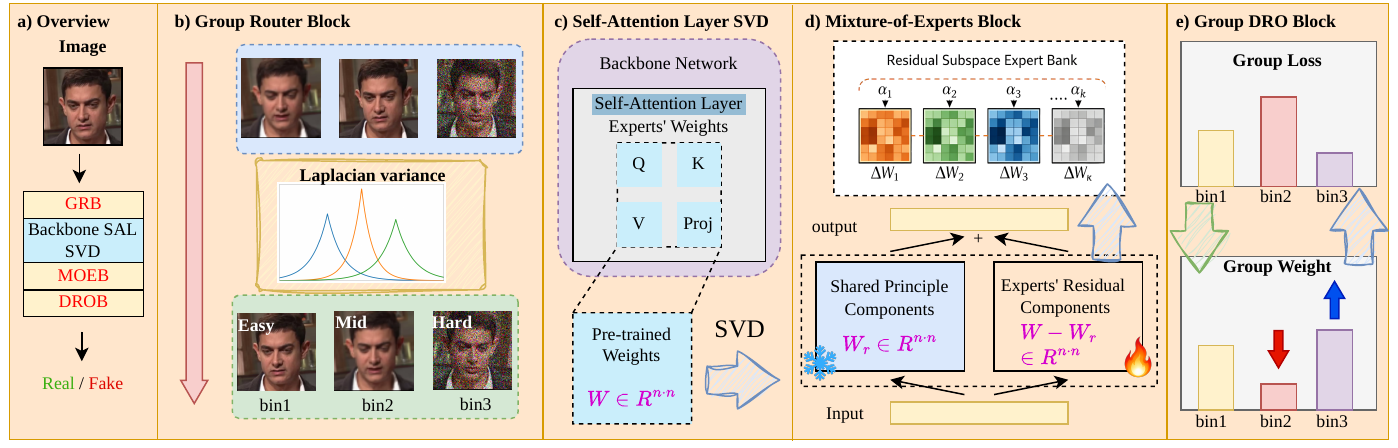}
    \caption{Overall pipeline of ACV Lab's \textit{Quality-Aware Multi-Expert Routing with Robust Optimization for Deepfake Detection}.}
    \label{fig:acv}
\end{figure*}

Team ACVLAB proposes a robust two-stage framework named Select and Detect, specifically designed for Quality-Aware Expert Routing and Robust Optimization in deepfake detection. The primary objective of this framework is to mitigate "shortcut learning," where detectors erroneously rely on image-quality artifacts rather than manipulation evidence.

The first stage, the "Select" phase, utilizes an Effort-style~\cite{yan2025orthogonalsubspacedecompositiongeneralizable} fine-tuned CLIP ViT-L/14~\cite{radford2021clip, dosovitskiy2021imageworth16x16words} backbone to extract high-level semantic features while preserving strong visual priors. To handle heterogeneous environmental degradations, the team introduced a Quality-Aware Multi-Expert Routing module. This module computes a lightweight image-quality proxy to estimate the quality-related grouping signal
 of the input, dynamically routing the features to specialized expert heads tailored for specific quality regimes. By isolating clean and degraded samples, the network avoids gradient interference and prevents performance collapse under heavy compression or noise.

The second stage, the "Detect" phase, focuses on Robust Optimization to ensure generalization across unseen distributions. Instead of standard empirical risk minimization, Team ACVLAB adopted Group Distributionally Robust Optimization (GroupDRO)~\cite{sagawa2019distributionally}. This strategy partitions the training data into quality-based groups and reweights the loss to minimize the risk of the worst-performing groups. Furthermore, a patch-level evidence aggregation scheme is implemented to capture localized manipulation traces that might be weakened by global degradation. For technical details,  \cref{fig:acv} provides a graphical representation of the Select and Detect pipeline.

%% file: sec/methods/reagvis_description.tex
\begin{figure*}
    \centering
    \includegraphics[width=0.85\linewidth]{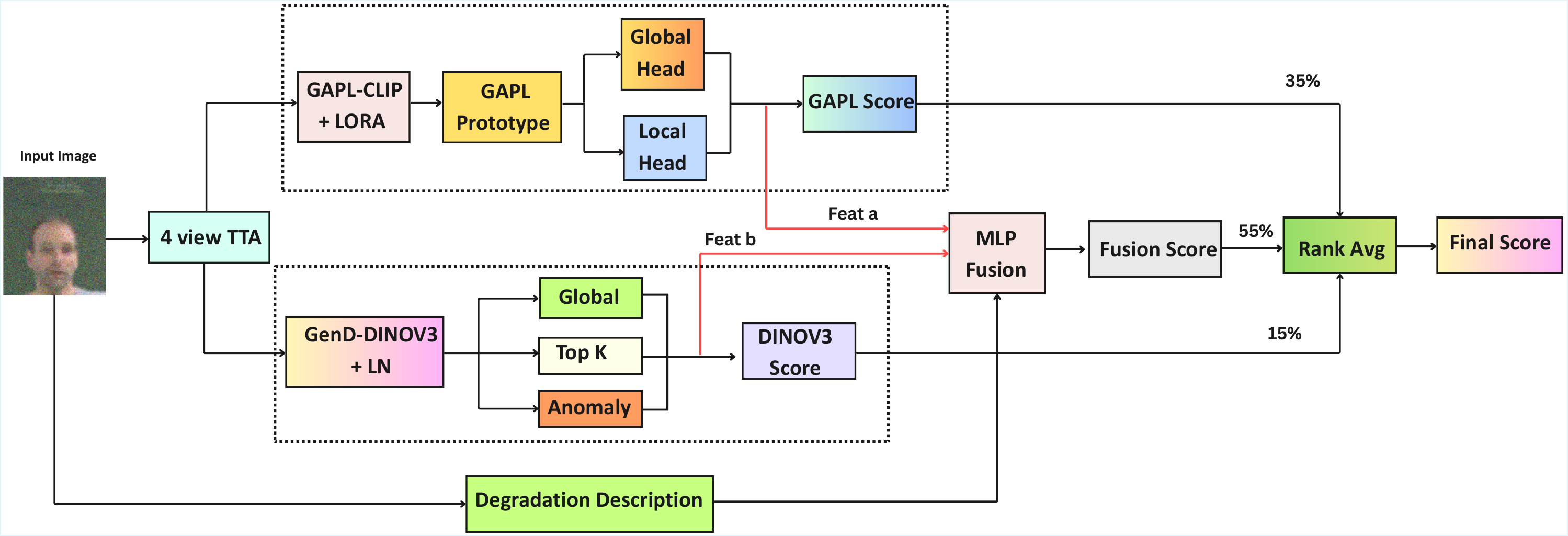}
    \caption{Overall pipeline of Reagvis Labs's \textit{Beyond Backbones: Degradation-Aware Prototype Fusion for Robust Deepfake Detection}.}
    \label{fig:reagvis}
    \vspace{-1mm}
\end{figure*}

Team REAGVIS develops a multi-backbone fusion framework for deepfake detection under severe image degradation. The pipeline has four components (\cref{fig:reagvis}): (1)~a CLIP-based backbone with generator-aware prototype learning (GAPL), (2)~a complementary DINOv3~\cite{simeoni2025dinov3} backbone with GenD~\cite{gend2024} deepfake-tuned initialization, (3)~a degradation-aware fusion MLP, and (4)~rank-based multi-model score calibration. The final AUC is \textbf{84.3} on the competition test set.

\textbf{Backbone A -- CLIP-GAPL:}
CLIP ViT-L/14~\cite{radford2021clip} fine-tuned with LoRA~\cite{Hu2021LoRALA} ($r{=}16$, $\alpha{=}32$, applied to $W_q, W_k, W_v$). The pooled CLS token ($\mathbb{R}^{1024}$) is projected to a 128-dim forensic space and matched against $K{=}64$ learnable GAPL prototypes~\cite{gapl2026} via multi-head cross-attention (4 heads), producing $z_{\text{global}} \in \mathbb{R}^{128}$. A local head applies soft-attention over 256 ViT patch tokens producing $z_{\text{local}} \in \mathbb{R}^{128}$. Standalone prediction:
\begin{equation}
    \ell_{\text{cg}} = 0.70 \cdot \ell_{\text{fuse}}([z_g; z_l]) + 0.15 \cdot \ell_{\text{global}} + 0.15 \cdot \ell_{\text{local}}
\end{equation}

\textbf{Backbone B -- DINOv3-GenD:}
DINOv3 ViT-Large (304M params)~\cite{oquab2024dinov2} initialized with GenD deepfake-tuned weights~\cite{gend2024}. Only LayerNorm parameters are tuned (${\sim}0.03\%$). Uses the same GAPL global head, plus an enhanced dual local head:
\begin{itemize}[nosep]
    \item \textbf{Top-$k$ head:} Scores 196 patches via MLP, selects top-32, softmax pooling $\to z_{h1} \in \mathbb{R}^{128}$
    \item \textbf{Anomaly head:} Per-patch anomaly as L2 distance from mean, top-32, learned pooling $\to z_{h2} \in \mathbb{R}^{128}$
\end{itemize}
Local output $z_{\text{local}} = [z_{h1}; z_{h2}] \in \mathbb{R}^{256}$, fused representation $[z_g; z_l] \in \mathbb{R}^{384}$. A detached degradation auxiliary head ($1024 \to 128 \to 5$) predicts degradation type during training only.

\textbf{Fusion MLP:}
Both backbones frozen. Features concatenated with 5 degradation descriptors (brightness, contrast, blur, grayscale flag, salt-and-pepper level):
$[z_g; z_l; \text{L2norm}(f_{\text{dino}}); d] \in \mathbb{R}^{1285} \xrightarrow{\text{LN+GELU+Drop(0.3)}} 256 \to 128 \to 1$.
The 363K-parameter MLP learns degradation-adaptive weighting.

\textbf{Rank-Based Score Calibration:}
4-view TTA (\{orig, hflip\} $\times$ \{center-crop, direct-resize\}), average logits pre-sigmoid, then:
\begin{equation}
    r_{\text{final}} = 0.55 \cdot \text{Rank}(\hat{p}_{\text{fusion}}) + 0.30 \cdot \text{Rank}(\hat{p}_{\text{cg}}) + 0.15 \cdot \text{Rank}(\hat{p}_{\text{dv3}})
\end{equation}

%% file: sec/methods/hit_description.tex
\begin{figure}
    \centering
    \includegraphics[width=\linewidth]{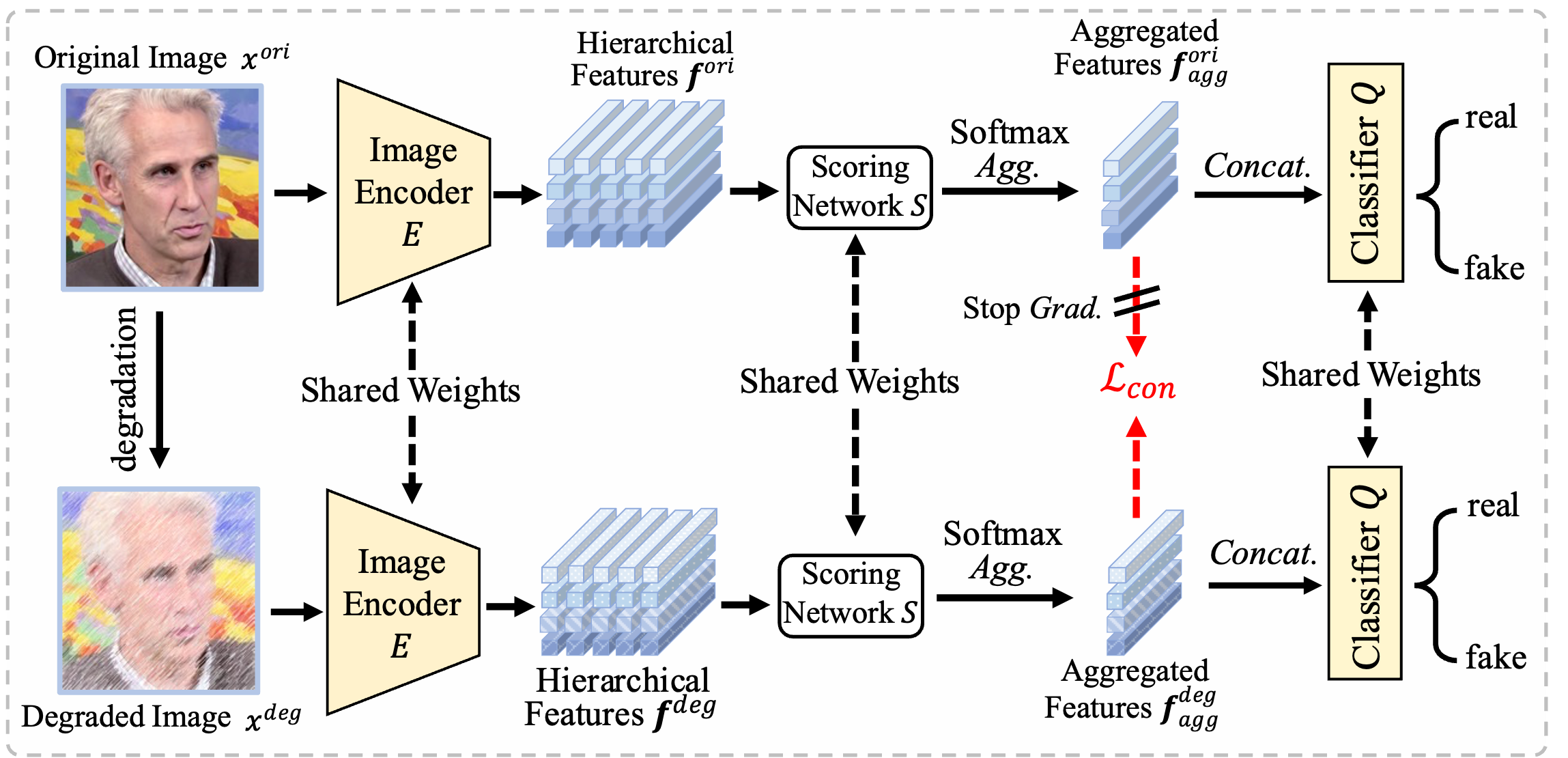}
    \caption{Overall pipeline of HIT-VIRLAB's \textit{Hierarchical Adaptive Feature Aggregation with Degraded-Original Consistency Learning for Robust Deepfake Detection}.}
    \label{fig:hit}
\end{figure}

Their solution aims to improve the robustness of deepfake detection under strong perturbations and real-world degradations (\cref{fig:hit}). Firstly, they construct a large-scale hybrid deepfake training dataset by collecting data from multiple publicly available sources, including FF++~\cite{roessler2019faceforensicspp}, DFDC~\cite{Dolhansky2020DFDC}, FakeAVCeleb~\cite{khalid2022fakeavcelebnovelaudiovideomultimodal}, Celeb-DF++~\cite{cdfv3}, DF40~\cite{yan2024df40}, and DDL \cite{miao2025ddllargescaledatasetsdeepfake}. The resulting million-scale dataset contains diverse forgery generation methods and visual conditions, providing rich variations for training robust deepfake detectors.

The framework adopts a Vision Transformer (ViT)~\cite{dosovitskiy2021imageworth16x16words} backbone initialized with FSFM ViT-B~\cite{wang2025fsfm} pretrained weights to extract features from input images. To better capture comprehensive forgery artifacts, they introduce a Hierarchical Adaptive Feature Aggregation (HAFA) module that leverages hierarchical features from multiple transformer layers. A learnable scoring network adaptively estimates token importance and aggregates informative tokens, enabling the model to integrate complementary cues ranging from low-level artifacts to high-level semantic information and emphasize critical forgery evidence.

Furthermore, they propose a Degraded-Original Consistency Learning (DOCL) strategy to improve robustness against common image degradations. During training, degraded samples are generated using a perturbation pipeline that simulates realistic distortions. A hierarchical consistency loss is applied to enforce feature consistency between original and degraded images. The final prediction is produced by an MLP classifier based on the concatenated aggregated hierarchical features.

%% file: sec/methods/anonymous1_description.tex
Their solution is based on parameter-efficient fine-tuning of the \texttt{vit\_large\_patch14\_clip\_224.openai} backbone, i.e., CLIP ViT-L/14~\cite{radford2021clip}, using LoRA~\cite{Hu2021LoRALA}. The training pipeline is built on the \textbf{OpenMMSec}~\cite{du2026openmmsec} dataset, which contains approximately \textbf{32.5K} images collected and reorganized from multiple public sources, including deepfake, AIGC, image manipulation, and document-forgery related datasets. To improve robustness, they organize the data into four domains: \textbf{AIGC}, \textbf{deepfake}, \textbf{doc}, and \textbf{imdl}.

During the competition, they explored multiple settings, including training with all domains jointly, training with only the deepfake domain, and fine-tuning with different LoRA ranks. Their final submission adopts a \textbf{two-stage} strategy. In the first stage, they pre-fine-tune the CLIP ViT-L/14 backbone on all domains of OpenMMSec. In the second stage, they continue fine-tuning on the \textbf{deepfake} domain only. During this second stage, they maintain a \textbf{1:1 real/fake ratio}; when the number of real images is insufficient, additional real images are randomly sampled from other domains. The final model uses \textbf{LoRA rank 64} and \textbf{LoRA alpha 128}.

At inference time, they apply a lightweight test-time augmentation (TTA) strategy~\cite{timofte2016seven} with three views: the original image, the horizontally flipped image, and the image rotated by $90^\circ$. The final prediction is obtained by averaging the three scores. This design improves robustness while maintaining practical efficiency.

%% file: sec/methods/zeke_description.tex
They adopt a deepfake detection approach based on a pre-trained CLIP ViT-Large (ViT-L/14)~\cite{dosovitskiy2021imageworth16x16words, radford2021clip} model released as part of the DF40 \cite{yan2024df40} benchmark. Rather than training or fine-tuning a model, they directly use the provided checkpoint for inference.

Their goal is to evaluate how well a large-scale vision-language model, trained on diverse data, generalizes to unseen manipulations and degradations present in the NTIRE challenge.

The pipeline operates purely on image inputs:
\begin{itemize}
\item Each input image is resized to $224 \times 224$
\item The image is passed through the CLIP ViT\cite{dosovitskiy2021imageworth16x16words} image encoder
\item A classification head outputs a prediction score representing the likelihood of the image being a deepfake
\end{itemize}

No temporal modeling or video-based aggregation is used. Each frame is processed independently, ensuring compatibility with the challenge constraints.

%% file: sec/methods/tcd_description.tex
TCD VISION uses the Effort~\cite{yan2025orthogonalsubspacedecompositiongeneralizable} detector (CLIP ViT-L/14~\cite{radford2021clip, dosovitskiy2021imageworth16x16words} with SVD-based parameter-efficient fine-tuning) from the DeepfakeBench~\cite{Yan2023DeepfakeBenchAC} framework, trained through a three-stage pipeline. Stage A pretrains on multiple public face-swap datasets for broad generalization. Stage B continues from Stage A with PMM-inspired heavy degradation augmentation to teach robustness under JPEG compression, blur, noise, resize, shadows, and overlay corruptions. Stage C fine-tunes on the competition data using 5-fold cross-validation with the same augmentation recipe. The final prediction averages fold outputs with horizontal-flip test-time augmentation~\cite{timofte2016seven}. The method prioritizes degradation robustness over architectural novelty, motivated by prior work showing that strong augmentation can substantially reduce the clean-to-degraded performance gap.

%% file: sec/methods/psu_description.tex
\textbf{PRISM} (\cref{fig:psu}) is a heterogeneous ensemble detector for AI-generated images, designed for robustness under JPEG compression, blur, noise, rescaling, and cropping. Their core hypothesis is that no single pre-training objective encodes the full forensic artefact manifold \cite{wang2020cnn,corvi2023diffusion}: contrastive objectives capture semantic inconsistencies; self-supervised patch objectives preserve texture discontinuities; supervised CNNs encode low-frequency spectral anomalies. Robust detection therefore requires \emph{explicit paradigm diversity}.

\noindent\textbf{Encoder pool:} They instantiate $K{=}7$ encoders $\{\phi_k\}_{k=1}^{K}$ across three paradigms (Table~\ref{tab:psu_suppl}). For vision-language (VL) encoders (CLIP \cite{radford2021clip}, SigLIP \cite{zhai2023siglip}, EVA02 \cite{fang2023eva02}), features are L2-normalised onto the unit hypersphere \cite{wang2020understanding}, preserving the contrastive geometry. DINOv2 \cite{oquab2024dinov2}, ConvNeXt \cite{liu2022convnext}, and EfficientNet-V2 \cite{Tan2019EfficientNetRM} remain frozen.

\noindent\textbf{Paradigm-aware tuning:} VL encoders undergo \emph{LayerNorm tuning} \cite{wortsman2022wiseft}: only LN scale/shift (${\approx}0.03\%$ of weights) are updated, preventing catastrophic forgetting while adapting internal normalisation to the forensic domain.

\noindent\textbf{Robust ensemble:} Each model is scored on held-out validation data under all degradation types, giving per-model robust AUC $A_k^\text{rob}$. Normalised weights $w_k {=} A_k^\text{rob}/\sum_j A_j^\text{rob}$ drive the ensemble. The final prediction with horizontal-flip TTA is:
\begin{equation}
  \hat{p}
  = \tfrac{1}{2}\!
    \left[
      \textstyle\sum_k w_k p_k(x)
      + \textstyle\sum_k w_k p_k(\mathcal{F}(x))
    \right].
  \label{eq:ensemble}
\end{equation}

%% file: sec/methods/ai4good_description.tex
They train binary classifiers exclusively on the 500 real images from the competition data using Self-Blended Images (SBI)~\cite{Shiohara2022SBI} combined with PMM degradation~\cite{hopf2025practicalmanipulationmodelrobust} (\cref{fig:ai4good}). No real fake images are used during training — pseudo-fakes are generated on-the-fly from pairs of real images. They build an ensemble of multiple models, including ConvNext~\cite{liu2022convnext}, DeiT~\cite{touvron2021trainingdataefficientimagetransformers}, and ViT~\cite{dosovitskiy2021imageworth16x16words}.

%% file: sec/methods/acube_description.tex
The architecture (\cref{fig:acube}) consists of two branches: an RGB backbone and a frequency branch. The RGB branch uses a ConvNeXt-Small~\cite{liu2022convnext} model pretrained on ImageNet~\cite{Hendrycks2019ImagenetPC} to extract semantic features, while the frequency branch processes the Fourier magnitude spectrum of the grayscale image to capture artifact-level inconsistencies.

Features from both branches are fused and passed through a lightweight classification head with Layer Normalization and dropout, ensuring good generalization on small datasets.

To improve robustness, moderate augmentations such as blur, compression, noise, color jitter, and geometric transformations are applied, while avoiding extreme degradations that harm frequency information. Class imbalance is handled using a WeightedRandomSampler.

The training strategy is tailored for small datasets. The backbone is initially frozen and later fine-tuned with a lower learning rate. Additional regularization includes label smoothing, mixup, weight decay, and dropout.

During validation, test-time augmentation (TTA) is applied using resized and flipped inputs. For inference, a lightweight TTA strategy~\cite{timofte2016seven} is used by averaging predictions from the original and horizontally flipped images.

Overall, the pipeline includes preprocessing, augmentation, dual-branch feature extraction (RGB + FFT), feature fusion, balanced training, and efficient inference with TTA. 

%% file: sec/methods/ntr_description.tex
NTR adopts a linear probing strategy on top of the frozen DINOv3 ViT-B/16 backbone \cite{simeoni2025dinov3}, Meta's latest vision foundation model pretrained on 1.69 billion image-text pairs. A single linear classification head is trained on the official challenge training set using 5-fold stratified cross-validation. At inference, they ensemble predictions from all five folds by averaging sigmoid probabilities, with horizontal-flip test-time augmentation (TTA)~\cite{timofte2016seven}.

The key motivation is that large-scale vision foundation models encode rich semantic and structural representations that transfer well to downstream forgery detection, even when only a lightweight linear head is trained. The frozen backbone prevents overfitting to the small training set while leveraging generalizable features learned from web-scale pretraining.

%% file: sec/4_conclusion.tex
\section{Conclusion}

This report presents the summary of the NTIRE 2026 Robust Deepfake Detection Challenge, including 14 finally submitted methods. The challenge's size, with $>300$ participants and $>50$ leaderboard submissions, underscores the importance and community interest in robust deepfake detection. Top-performing methods generally leverage large pretrained foundation models, often ensembles and degradation models.  This shows that high-quality pretraining can prevent overfitting, while exposure to low-quality images during training improves robustness.

\section*{Acknowledgments}
This work was partially supported by the Humboldt Foundation. We thank the NTIRE 2026 sponsors: OPPO, Kuaishou, and the University of Wurzburg (Computer Vision Lab).

%% file: sec/X_suppl.tex
\clearpage
\appendix
\setcounter{page}{1}
\maketitlesupplementary

\section{Teams}
\label{sec:teams}

\input{sec/methods/shallowreal_team}
\input{sec/methods/intsig_team}
\input{sec/methods/antint_team}
\input{sec/methods/hcmus_team}
\input{sec/methods/acvlab_team}
\input{sec/methods/reagvis_team}
\input{sec/methods/hit_team}
\input{sec/methods/anonymous1_team}
\input{sec/methods/zeke_team}
\input{sec/methods/tcd_team}
\input{sec/methods/psu_team}
\input{sec/methods/ai4good_team}
\input{sec/methods/acube_team}
\input{sec/methods/ntr_team}

\section{Method comparison}
\Cref{tab:methodinfos} shows a comparative overview of the submitted methods.

\begin{table*}[bt]
    \centering
    \begin{tabular}{c:c|c|c|c|c}
        \toprule
        Rank & \textbf{Name} & Input & Extra data & Base Model & \# Params.  \\
        \midrule
        1 & \textbf{ShallowReal} & (656, 656, 3) & Yes & DINOv3-L & 600M \\
        2 & \textbf{INTSIG} & multiple & Yes & 4x DINOv3-H + MetaCLIP-H & 4.2B \\
        3 & \textbf{AntInternational} & (384, 384, 3) & Yes & DINOv3-ViT & 7B \\
        4 & \textbf{HCMUS-Aqua} & (256, 256, 3) & Yes & DINOv2-G + CLIP-L & ~1.5B \\
        5 & \textbf{ACVLab} & (224, 224, 3) & Yes & CLIP ViT-L/14 &  505M \\
        6 & \textbf{Reagvis Labs} & (224, 224, 3) & SBI & CLIP-L + DINOv3-L & 732M \\
        7 & \textbf{HIT-VIRLAB} & (224, 224, 3) & Yes & Fsfm ViT-B [8] & 92M \\
        8 & \textbf{\textit{Anonymous}} & (224, 224, 3) & Yes & CLIP ViT-L/14 & 437M \\
        9 & \textbf{Zeke} & (224, 224, 3) & No & DF40-CLIP-L & 430M \\
        10 & \textbf{TCD Vision} & (224, 224, 3) & Yes & Effort & 303M \\
        11 & \textbf{PSU Team} & (224, 224, 3) & Yes & multiple & 2B \\
        12 & \textbf{AI4Good} & (224, 224, 3) & SBI & multiple & 430M \\
        13 & \textbf{Acube} & (224, 224, 3) & No & ConvNeXt-Small + FFT & 50M \\
        14 & \textbf{NTR} & (224, 224, 3) & No & DINOv3 ViT-B/16 & 86M\\
        \bottomrule
    \end{tabular}
    \caption{Overview of method specification.}
    \label{tab:methodinfos}
\end{table*}

\section{Additional method details}

In \cref{tab:psu_suppl}, \cref{fig:psu}, \cref{fig:ai4good}, and \cref{fig:acube}, we show some details that did not fit into the main paper.

\input{sec/methods/psu_suppl}
\input{sec/methods/ai4good_suppl}
\input{sec/methods/acube_suppl}

%% file: sec/methods/shallowreal_team.tex
\subsection*{ShallowReal}
\noindent\textit{\textbf{Title: }} DINO-MAC\\
\noindent\textit{\textbf{Members:}}\\ 
\textit{Chenfan Qu$^1$(\href{mailto:hongge568@126.com}{hongge568@126.com})}, Junchi Li$^{2}$\\
\noindent\textit{\textbf{Affiliations: }}\\
$^1$ South China University of Technology 1 \\
$^2$ Zhejiang University 2 \\

%% file: sec/methods/intsig_team.tex
\subsection*{INTSIG}
\noindent\textit{\textbf{Title: }} LOGER: Local-Global Ensemble for Robust Deepfake Detection in the Wild\\
\noindent\textit{\textbf{Members:}}\\
\textit{Fei Wu$^{1,2}$(\href{mailto:wu_fei@sjtu.edu.cn}{wu\_fei@sjtu.edu.cn})}, \textit{Dagong Lu$^{1}$(\href{mailto:dagong_lu@intsig.net}{dagong\_lu@intsig.net})}, \textit{Mufeng Yao$^1$(\href{mailto:mufeng_yao@intsig.net}{mufeng\_yao@intsig.net})}, \textit{Xinlei Xu$^1$(\href{mailto:xinlei_xu@intsig.net}{xinlei\_xu@intsig.net})}, \textit{Fengjun Guo$^1$(\href{mailto:fengjun_guo@intsig.net}{fengjun\_guo@intsig.net})}\\
\noindent\textit{\textbf{Affiliations: }}\\
$^1$ INTSIG, Shanghai, China \\
$^2$ Shanghai Jiao Tong University, Shanghai, China \\

%% file: sec/methods/antint_team.tex
\subsection*{Ant International}
\noindent\textit{\textbf{Title: }} An Ensemble of Architecturally-Diverse Large-Scale Vision Transformers\\
\noindent\textit{\textbf{Members:}}\\ 
\textit{Yongwei Tang$^1$(\href{mailto:tangyongwei.tyw@ant-intl.com}{tangyongwei.tyw@ant-intl.com})}, \textit{Zhiqiang Yang$^2$(\href{mailto:yangzhiqiang.yzq@antgroup.com}{yangzhiqiang.yzq@antgroup.com})}, \textit{Zhiqiang Wu$^1$(\href{mailto:hanli.wzq@ant-intl.com}{hanli.wzq@ant-intl.com})}, \textit{Jia Wen Seow$^1$(\href{mailto:jiawen.seow@ant-intl.com}{jiawen.seow@ant-intl.com})}, \newline
\textit{Hong Vin Koay$^1$(\href{mailto:hongvin.koay@ant-intl.com}{hongvin.koay@ant-intl.com})}, \textit{Haodong Ren$^1$(\href{mailto:renhaodong.rhd@ant-intl.com}{renhaodong.rhd@ant-intl.com})}, \newline \textit{Feng Xu$^1$(\href{mailto:fuyu.xf@ant-intl.com}{fuyu.xf@ant-intl.com})},
\textit{Shuai Chen$^1$(\href{mailto:shuai.cs@ant-intl.com}{shuai.cs@ant-intl.com})}\\
 \newline
\noindent\textit{\textbf{Affiliations: }}\\
$^1$ Ant International \\
$^2$ Ant Group \\

%% file: sec/methods/hcmus_team.tex
\subsection*{HCMUS-Aqua}
\noindent\textit{\textbf{Title: }} Robust Deepfake Detection via Multi-Stream DINO-CLIP Fusion and Discretized Voting\\
\noindent\textit{\textbf{Members:}}\\ 
\textit{Minh-Khoa Le-Phan$^1$(\href{mailto:lpmkhoa22@apcs.fitus.edu.vn}{lpmkhoa22@apcs.fitus.edu.vn})}, Minh-Hoang Le$^{1}$, 
Trong-Le Do$^1$, Minh-Triet Tran$^1$\\
\noindent\textit{\textbf{Affiliations: }}\\
$^1$ University of Science, VNU-HCM. Vietnam National University, Ho Chi Minh City, Vietnam \\

%% file: sec/methods/acvlab_team.tex
\subsection*{TEAM ACVLAB}
\noindent\textit{\textbf{Title:}} Select and Detect: Quality-Aware Expert Routing and Robust Optimization for Deepfake Detection\\
\textit{\textbf{Members:}}\\
\textit{Chih-Yu Jian (\href{mailto:ru0354m3@gmail.com}{ru0354m3@gmail.com})}, Yi-Fan Wang, Bang-Kang Chen, You-Chen Chao,  Chia-Ming Lee, Fu-En Yang, Yu-Chiang Frank Wang, Chih-Chung Hsu\\
\textit{\textbf{Affiliations:}}\\
Institute of Intelligent Systems, National Yang Ming Chiao Tung University\\
Institute of Data Science, National Cheng Kung University\\
Department of Computer Science, University at Albany -- SUNY\\
NVIDIA, Taipei, Taiwan

%% file: sec/methods/reagvis_team.tex
\subsection*{Reagvis Labs}
\noindent\textit{\textbf{Title: }} Beyond Backbones: Degradation-Aware Prototype Fusion for Robust Deepfake Detection\\
\noindent\textit{\textbf{Members:}}\\
\textit{Praful Hambarde$^1$(\href{mailto:praful@iitmandi.ac.in}{praful@iitmandi.ac.in})}, Aashish Negi$^{1}$, Hardik Sharma$^{1}$,
Prateek Shaily$^2$, Jayant Kumar$^2$, Sachin Chaudhary$^2$, Akshay Dudhane$^3$, Amit Shukla$^1$\\
\noindent\textit{\textbf{Affiliations: }}\\
$^1$ Indian Institute of Technology Mandi \\
$^2$ University of Petroleum and Energy Studies (UPES) \\
$^3$ Mohamed bin Zayed University of Artificial Intelligence (MBZUAI) \\

%% file: sec/methods/hit_team.tex
\subsection*{HIT-VIRLAB}
\noindent\textit{\textbf{Title:}} Hierarchical Adaptive Feature Aggregation with Degraded-Original Consistency Learning for Robust Deepfake Detection\\
\noindent\textit{\textbf{Members:}}\\
\textit{Jielun Peng(\href{mailto:jielunpeng_hit@163.com}{{jielunpeng\_hit@163.com}})}, Yabin Wang,
Yaqi Li, jincheng Liu, Xiaopeng Hong\\
\noindent\textit{\textbf{Affiliations: }}\\
Harbin Institute of Technology \\

%% file: sec/methods/anonymous1_team.tex
\subsection*{Anonymous}
\noindent\textit{\textbf{Title: }} LoRA Fine-Tuning for CLIP

\noindent\textit{\textbf{Members:}} - 

\noindent\textit{\textbf{Affiliations: }} - 

%% file: sec/methods/zeke_team.tex
\subsection*{Zeke}
\noindent\textit{\textbf{Title: }} Robust Deepfake Detection using Large Scale Vision Transformers\\

\noindent\textit{\textbf{Members:}}\\ 
\textit{Krish Wadhwani$^1$(\href{mailto:wadhwan5@msu.edu}{wadhwan5@msu.edu}), Liam Fitzpatrick$^1$(\href{mailto:fitzp226@msu.edu}{fitzp226@msu.edu)}}\\

\noindent\textit{\textbf{Affiliations: }}\\
$^1$ Michigan State University \\

%% file: sec/methods/tcd_team.tex
\subsection*{TCD Vision}
\noindent\textit{\textbf{Title: }} Robust Deepfake Detection with Parameter-Efficient CLIP Fine-Tuning and PMM-Style Degradation Augmentation\\
\noindent\textit{\textbf{Members:}}\\ 
\textit{Utkarsh Tiwari$^1$(\href{mailto:tiwariu@tcd.ie}{tiwariu@tcd.ie})}\\
\noindent\textit{\textbf{Affiliations: }}\\
$^1$ Trinity College Dublin \\

%% file: sec/methods/psu_team.tex
\subsection*{PSU TEAM}
\noindent\textit{\textbf{Title: }} PRISM: Paradigm-diverse Representation Integration for Synthesis-artifact Manifold Detection\\
\noindent\textit{\textbf{Members:}}\\ 
\textit{Bilel Benjdira$^1$(\href{mailto:bbenjdira@psu.edu.sa}{bbenjdira@psu.edu.sa})}, Anas M. Ali$^1$, Wadii Boulila$^1$\\
\noindent\textit{\textbf{Affiliations: }}\\
$^1$ Robotics and Internet-of-Things Laboratory, Prince Sultan University, Riyadh 12435, Saudi Arabia \\

%% file: sec/methods/ai4good_team.tex
\subsection*{AI4GOOD}
\noindent\textit{\textbf{Title:}} Self-Supervised Adversarial Training for Robust Deepfake
Detection\\
\textit{\textbf{Members:}}\\
\textit{Cristian Lazo Quispe (\href{mailto:clazoq@uni.pe}{ru0354m3@gmail.com})}\\
\textit{\textbf{Affiliations:}}\\
Universidad Nacional de Ingenier´ıa (UNI), Lima, Per´u

%% file: sec/methods/acube_team.tex
\subsection*{ACUBE}

\noindent\textit{\textbf{Title: }} Robust Deepfake Detection using ConvNext\\
\noindent\textit{\textbf{Members:}}\\ 
\textit{Aishwarya A(\href{mailto:aishwaryashyamala14@gmail.com}{aishwaryashyamala14@gmail.com})}, Akshara S, 
Ashwathi N\\
\textit{\textbf{Affiliations:}}\\
Department of Artificial Intelligence and Data Science, Shiv Nadar University
Chennai

%% file: sec/methods/ntr_team.tex
\subsection*{NTR}
\noindent\textit{\textbf{Title:}} DINOv3 ViT-B/16 Linear Probe Ensemble for Robust Deepfake Detection\\
\noindent\textit{\textbf{Members:}}\\
\textit{Jiachen Tu$^1$ (\href{mailto:jtu9@illinois.edu}{jtu9@illinois.edu})},
Guoyi Xu$^1$, Yaoxin Jiang$^1$, Jiajia Liu$^1$, Yaokun Shi$^1$\\
\noindent\textit{\textbf{Affiliations:}}\\
$^1$ University of Illinois Urbana-Champaign\\

%% file: sec/methods/psu_suppl.tex
\begin{figure*}
    \centering
    \includegraphics[width=\linewidth]{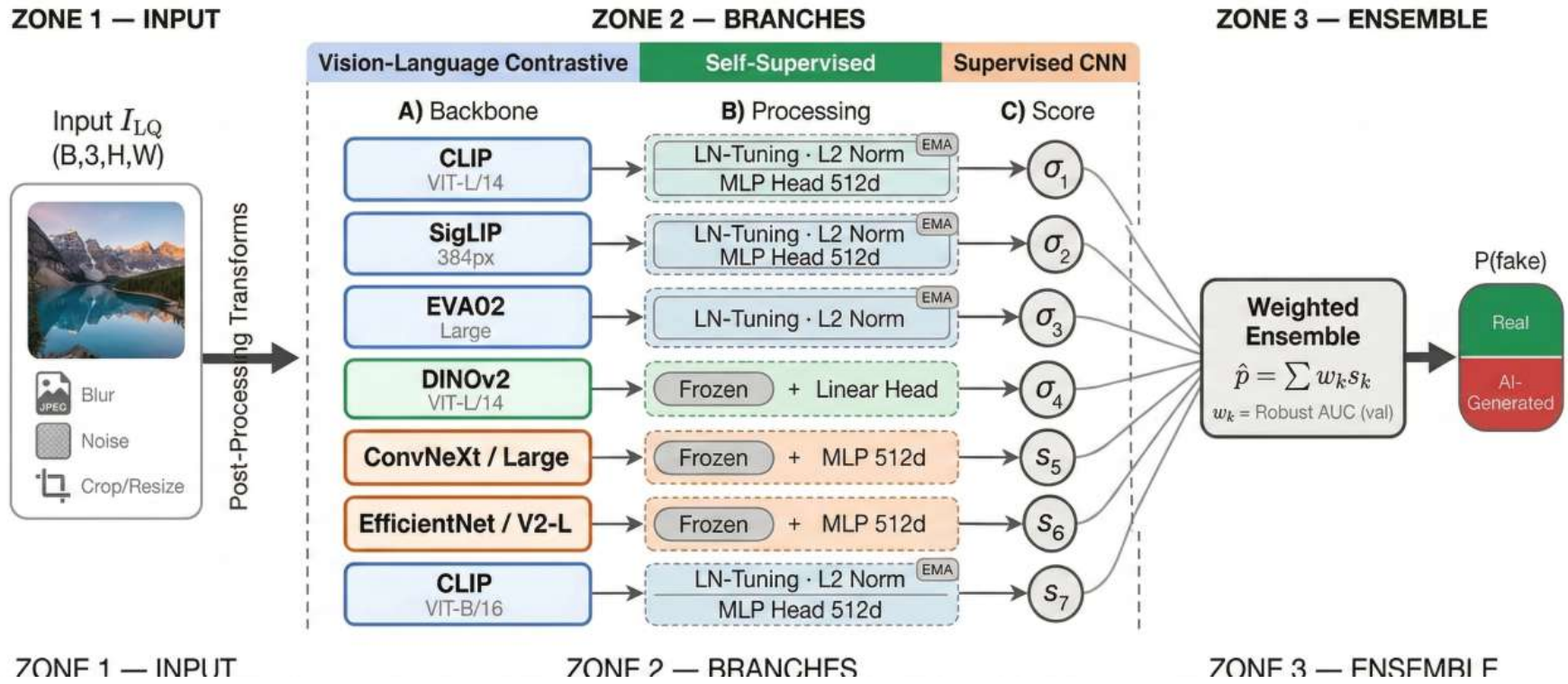}
    \caption{Overall pipeline of PSU's \textit{PRISM: Paradigm-diverse Representation Integration for Synthesis-artifact Manifold Detection}.}
    \label{fig:psu}
\end{figure*}
\begin{table}
  \centering
  \caption{PRISM encoder pool. $d$: feature dim. LN = LayerNorm-only; F = frozen.}
  \label{tab:psu_suppl}
  \footnotesize
  \setlength{\tabcolsep}{3.5pt}
  \begin{tabular}{llccc}
    \toprule
    Encoder           & Paradigm        & $d$  & Tune & Head    \\
    \midrule
    CLIP ViT-L/14     & VL-Contrastive  & 1024 & LN   & MLP-512 \\
    SigLIP ViT-L/16   & VL-Contrastive  & 1024 & LN   & MLP-512 \\
    EVA02-Large       & VL-Contrastive  & 1024 & LN   & MLP-512 \\
    DINOv2 ViT-L/14   & Self-Supervised & 1024 & F    & Linear  \\
    ConvNeXt-Large    & Supervised CNN  & 1536 & F    & MLP-512 \\
    EfficientNet-V2-L & Supervised CNN  & 1280 & F    & MLP-512 \\
    CLIP ViT-B/16     & VL-Contrastive  &  512 & LN   & MLP-256 \\
    \bottomrule
  \end{tabular}
\end{table}

%% file: sec/methods/ai4good_suppl.tex
\begin{figure*}
    \centering
    \includegraphics[width=\linewidth]{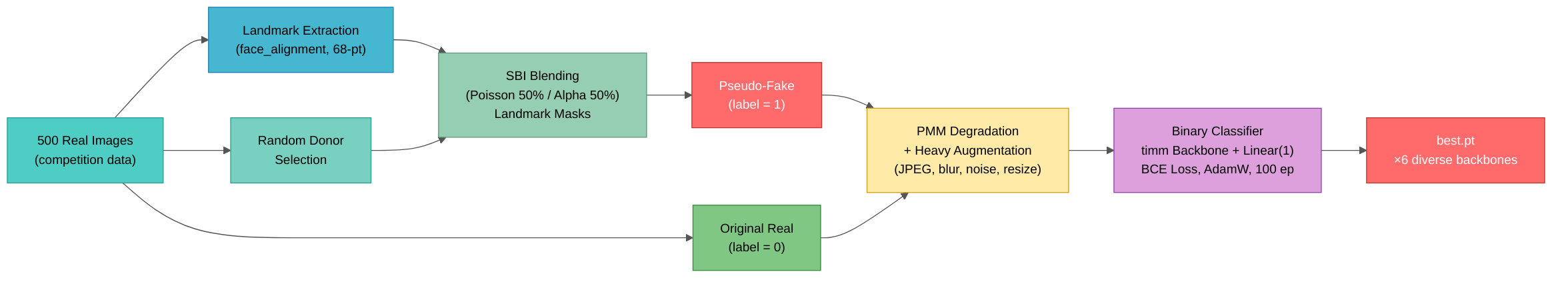}
    \caption{Overall pipeline of AI4Good's \textit{Self-Supervised Adversarial Training for Robust Deepfake
Detection}.}
    \label{fig:ai4good}
\end{figure*}

%% file: sec/methods/acube_suppl.tex
\begin{figure}
    \centering
    \includegraphics[width=\linewidth]{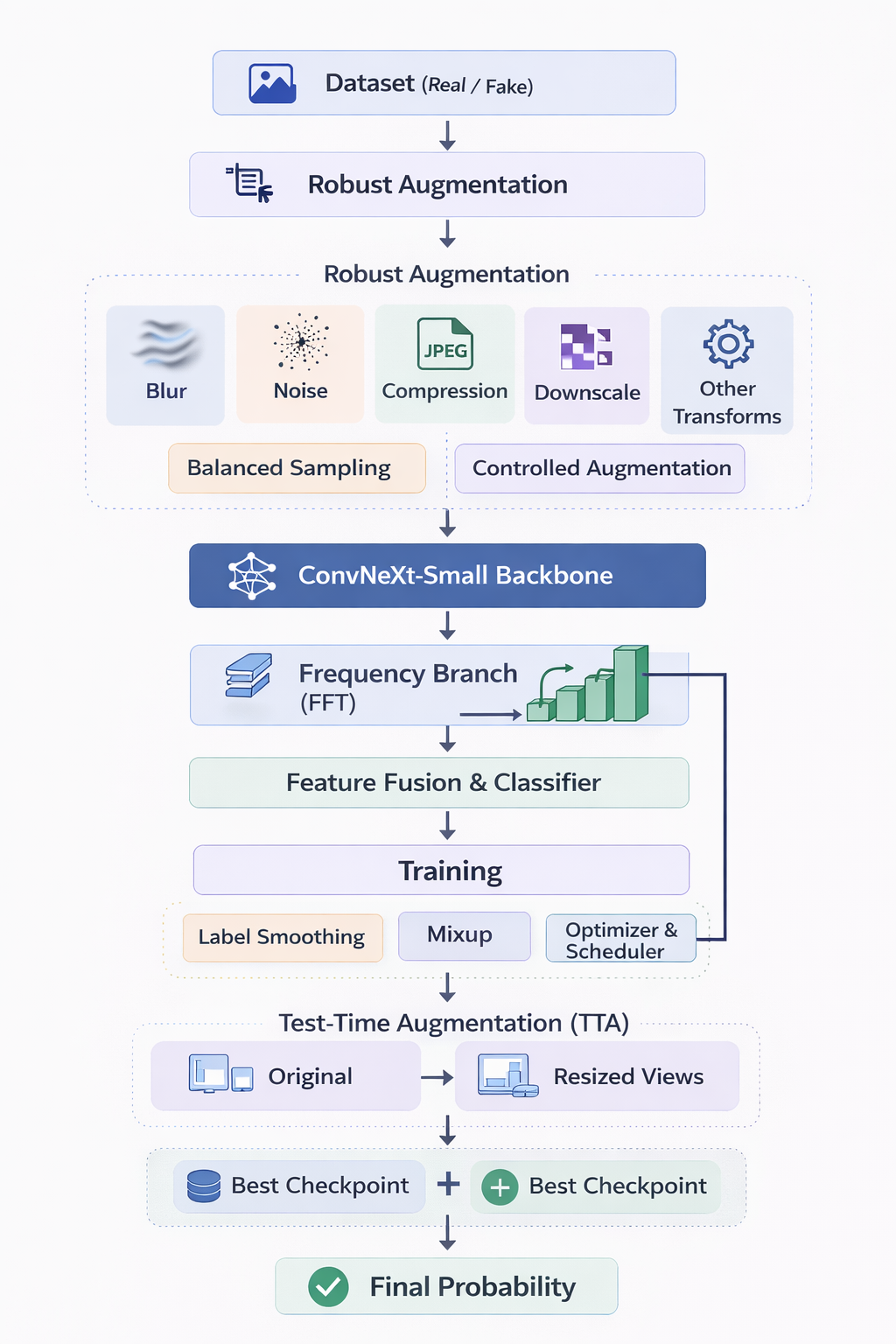}
    \caption{Overall pipeline of ACUBE's \textit{Robust Deepfake Detection using ConvNeXt with Frequency-Aware Fusion and Regularized Training Strategy}.}
    \label{fig:acube}
\end{figure}